\documentclass{Styles/svproc}

\usepackage{amsfonts}
\usepackage{amsmath}
\usepackage{graphicx}
\usepackage{url}
\usepackage{xcolor}
\usepackage[hidelinks]{hyperref}
\usepackage{algorithm}
\usepackage{algpseudocode}
\usepackage{bm}

\newcommand{\q}{\mathbf{q}}
\newcommand{\dq}{\dot{\mathbf{q}}}
\newcommand{\uin}{\mathbf{u}}
\newcommand{\R}{\in\mathbb{R}}
\newcommand{\traj}{\q_{des}(t)}
\newcommand{\trajs}{\q_{des}(s(t))}
\newcommand{\dtraj}{\dot{\q}_{des}(t)}
\newcommand{\dtrajs}{\q'_{des}(s(t))}
\newcommand{\dqs}{\q'(s)}

\newcommand{\myparagraph}[1]{\paragraph{#1}\mbox{} \vspace{\baselineskip}}

\begin{document}

    \mainmatter
    
    \title{Safe Multimodal Communication in Human-Robot Collaboration}
    \titlerunning{Safe Multimodal Communication}
    \author{Davide Ferrari \and Andrea Pupa \and Alberto Signoretti \and Cristian Secchi}
    \authorrunning{Davide Ferrari et al.} 
    
    \tocauthor{Davide Ferrari, Andrea Pupa, Alberto Signoretti and Cristian Secchi}
    
    \institute{University of Modena and Reggio Emilia, Italy\\
    \email{\{davide.ferrari95, andrea.pupa,
        cristian.secchi\}@unimore.it}}
    
    \maketitle
    
    \begin{abstract}
    The new industrial settings are characterized by the presence of human and robots that work in close proximity, cooperating in performing the required job. Such a collaboration, however, requires to pay attention to many aspects. Firstly, it is crucial to enable a communication between this two actors that is natural and efficient. Secondly, the robot behavior must always be compliant with the safety regulations, ensuring always a safe collaboration.
    In this paper, we propose a framework that enables multi-channel communication between humans and robots by leveraging multimodal fusion of voice and gesture commands while always respecting safety regulations. The framework is validated through a comparative experiment, demonstrating that, thanks to multimodal communication, the robot can extract valuable information for performing the required task and additionally, with the safety layer, the robot can scale its speed to ensure the operator's safety.
    
    
    \end{abstract}

	\section{Introduction}\label{sec:introduction}

    Effective communication between humans and robots is a crucial element in collaborative robotics. As robots become increasingly present in work and home environments, the ability to communicate naturally and efficiently with humans becomes a determining factor for the success of the interaction and the achievement of common goals.
    \noindent
    According to literature \cite{REGENBOGEN2012NeuroImage} \cite{HOLLER2019TCS}, human communication is based on the coexistence and fusion of multiple different communicative modes (or channels), leading to the realization of a \textbf{multimodal communication model}. These modes can include verbal language, body language, gestures, facial expressions, and even the use of lights or sounds. In the context of collaborative robotics, implementing multimodal human-robot communication refers to the robot's ability to use different sensory modes to perceive, interpret, and generate communicative signals. This allows the robot to acquire more complete and accurate information about the intentions, desires, and emotions of the human it interacts with, facilitating a deeper understanding and an appropriate response to the needs of the human interlocutor.
    \noindent
    However, multimodal communication alone is not sufficient to ensure accurate understanding and appropriate response in human-robot collaboration (HRC) \cite{BAUER2008IJHR};
    this is where multimodal fusion comes into play \cite{PORIA2017InformationFusion}.
    \textbf{Multimodal fusion} is the process by which information from different sensory modalities is combined and integrated to obtain a coherent and complete representation of the surrounding environment and interactions with humans. This fusion of information allows the robot to benefit from the different perspectives provided by each modality, improving its understanding of human intentions and enabling a more precise and adaptable response. For example let's consider a situation where a collaborative robot is working in tandem with a human operator in a manufacturing company. During the interaction, the robot may use voice recognition to understand verbal instructions from the human operator, but simultaneously it can also monitor the operator's body language and facial expressions to detect any signs of stress or dissatisfaction. The fusion of information from these different sensory modalities allows the robot to have a more comprehensive understanding of the operator's intentions and emotions, enhancing its ability to provide an appropriate response and adapt to the needs of the human interlocutor. 
    In \cite{liu2018IEEE}, a deep learning-based multimodal fusion architecture for robust multimodal HRC manufacturing systems is proposed. Experiments using speech command, hand motion, and body motion, show that the proposed multimodal fusion model outperforms the three unimodal models. Other works \cite{mohd2022multimodal} have proposed multiple combinations of unimodal input and fusion architectures, focusing solely on improving the robot's communication capabilities, but lacking consideration for the safety of the operator.
    \noindent
    In conclusion, it is evident that multimodal human-robot communication and multimodal fusion play a central role in collaborative robotics; however, \textbf{safety} remains a crucial aspect to consider during human-robot interactions.
    In particular \cite{MERCKAERT2022} introduces a computationally efficient control scheme for safe human–robot interaction based on the Explicit-Reference-Governor formalism, ensuring that the robot can safely works close proximity to humans. In \cite{Palleschi2021IEEE} the authors proposed a trajectory planning algorithm to produce safe minimum-time trajectories in a shared workspace with humans, with the addition of a re-planning module to optimally adapt the generated trajectory online to ensure safety limits.
    In this paper we integrate a multimodal fusion architecture with a two layer framework, proposed in \cite{pupa2021safety}, that plans a trajectory ensuring a collision-free path by monitoring the skeleton of the operator. Our goal is to \textit{develop an architecture that can ensure natural and efficient multimodal human-robot communication while also ensuring safety} during operations by utilizing communication channels to seek input from the operator on how to resolve potential errors or unforeseen circumstances.
    
    \noindent
    This combination of multimodal communication and integrated safety represents a significant step towards advanced and efficient collaborative robotics, where the collaboration between humans and robots occurs harmoniously and safely. We hope that this work can provide a solid foundation for future research and developments in the field of collaborative robotics, leading to new solutions that improve the effectiveness, interaction, and safety of human-robot systems.


    \newpage
    
    \vspace{\baselineskip} \noindent
    Thus, the contributions of this paper are:
    \begin{itemize}
        \item A Multimodal Fusion Architecture using 3D Gestures and Voice.
        \item The integration of the Fusion with a Safety Layer to ensure the respect of the safety measures.
        \item An experimental validation by comparing the safe and unsafe architectures during a pick-and-place job.
    \end{itemize}
    \noindent
    This paper is organized in the following way: in Section II, we introduce the problem statement. In Sections III and IV, we describe the architecture, focussing on the communication channels and the safety layer. In Section V, we present the experimental validation, providing detailed information about the implementation, and analyze the obtained results. We conclude and outline some ideas for future works in Section VI.
    
    

    \section{Problem Statement}
    

    Consider a scenario of Human-Robot Collaboration in which a 6-dof velocity-controlled manipulator needs to collaborate and communicate with a human operator to fulfill a shared objective.

    \vspace{\baselineskip}\noindent
    The collaborative robot can be modeled as:
    \begin{equation}
        \dq = \uin,
        \label{eq:model}
    \end{equation}
    \noindent
    where $\dq\R^n$ represent the joints velocities and  $\uin\R^n$ the controller input.

    \vspace{\baselineskip}\noindent
    We consider scenarios where the human operator actively lead the collaboration, who is the most expert member. This is because we want to reproduce a situation in which the robot has to assists the user in performing a series of tasks, by following its instructions and replying to its questions. To this aim, the collaborative scenario is endowed with a set of sensors that enable the communication between the two actors implementing a communication strategy that can guarantee a natural and efficient exchange of information, in order to achieve good performance and approach the results of human-human collaboration. Multimodal fusion allows combining information from multiple communication channels, including voice and gestures, which represent the foundations of human communication since early communication development \cite{BURKHARDTREED2021}.

    \vspace{\baselineskip}\noindent
    Each communication command can be modeled as a desired final configuration $\q_{des}(t_f)=\q_f\R^n$ that the robot has to reach by executing safe trajectories $\traj\R^n$ from an initial configuration $\q_{des}(t_i)=\q_i\R^n$, ensuring compliance with ISO/TS 15066, which imposes a limit on the maximum speed in the direction of the operator \cite{pupa2021safety}:
    \begin{equation}
        \begin{split}
            v_{rh}(t)\le &\sqrt{v_h(t)^2+(a_{max}T_r)^2-2a_{max}(C + Z_d + Z_r-S_p(t))}+\\
            &-a_{max}T_r -v_h(t),
            \label{eq:vellimit}
        \end{split}
    \end{equation}

    \noindent
    where $v_{{rh}}(t)\in \mathbb{R}$ represents the velocity of the robot towards the human and $v_{h}(t)\in \mathbb{R}$ represents the velocity of the human. $a_{max} \in \mathbb{R}$ and $T_r\in\mathbb{R}$ are the maximum deceleration and the robot reaction time, respectively.

    \vspace{\baselineskip}\noindent
    In order to ensure compliance with the safety standards while keeping the overall path, we can explicitly isolate the magnitude of the velocity along the trajectory by applying a path-velocity decomposition and act on the derivative $\dot{s}$ of the curvilinear abscissa $s$ that parameterizes the geometric path $\trajs$:
    \begin{align}
        & &\traj &= \trajs &&t\in \begin{bmatrix}t_i, t_f\end{bmatrix},
        \label{eq:q_s}
        \\
        & &\dtraj &= \dtrajs\dot{s} &&t\in \begin{bmatrix}t_i, t_f\end{bmatrix},    
        \label{eq:q_s_dot}
    \end{align}


    \vspace{\baselineskip}\noindent
    In this paper we propose a multimodal fusion architecture that:
    \begin{itemize}
        \item Enables natural and efficient communication by utilizing two of the main channels of human communication.
        \item Performs multimodal fusion to combine information from multiple communication modes and extract an overall meaning from them.
        \item Ensures safety compliance during human-robot collaboration.
    \end{itemize}


    \newpage
\section{Proposed Architecture}

\begin{figure}[htbp]
        \centering
        \includegraphics[trim={5.5cm 6cm 5.5cm 7.5cm}, clip, width=12cm]{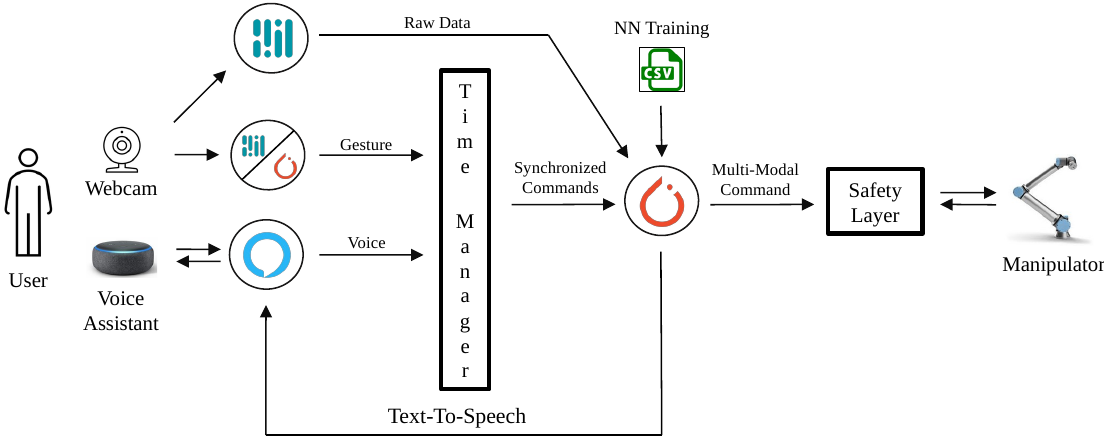}
        \caption{Multimodal Architecture}
        \label{fig:multimodal architecture}
    \end{figure}

    The proposed architecture, summarized in Figure \ref{fig:multimodal architecture}, consists of a \textbf{vocal communication} and a \textbf{gesture recognition} channels, which are then fused by a \textbf{multimodal fusion} algorithm, enabling bidirectional and dynamic communication between humans and robots. The information that comes from these two channels are firstly collected by a \textbf{time manager}, which is responsible for synchronizing and merging them into a single tensor, and subsequently fused with a \textbf{neural classifier} to obtain a coherent and comprehensive representation of the communicated message. Additionally, a text-to-speech channel has been added to let the robot able to provide information to the operator, such as feedback on the status of the task or the occurrence of errors or issues. The commands obtained from multimodal fusion are sent to the safety layer, which is responsible for planning and executing a trajectory while ensuring compliance with safety distances from the operator.
    
    \noindent
    The vocal communication channel, built around a commercial voice assistant, has been created through a custom application that connects a front-end interface for operator interaction with a back-end running locally to exchange information with the rest of the architecture. Additionally, a Text-To-Speech channel has been added to enable vocal feedback, allowing the robot to communicate with the operator.

    \vspace{\baselineskip} \noindent
    The gesture communication channel was created using a gesture recognition algorithm based on a neural network classifier. 
    A real-time video stream is captured by a webcam and each frame is processed frame-by-frame by a skeletonization algorithm that extract a series of keypoints containing the spatial coordinates of skeleton, face and hands. These keypoints are then encoded into a tensor representing a 3D gesture (a gesture that extends over time) and passed through the neural classifier to categorize the performed gesture. The classifier consists of an LSTM (Long Short-term Memory) layer \cite{Hochreiter1997NeuralComputation}, followed by several fully connected layers. It provides an output vector of probabilities indicating which of the trained gestures is most likely to have been executed.
    In addition, we have developed some ”raw functions”, which are functions that enhance the meaning of certain gestures. For example, the ”Point At” gesture requires the direction in which the user is pointing to provide meaningful information. When this gesture is recognized by the neural network, a specific function is triggered to calculate the direction by tracing a line intersecting shoulder, elbow, and wrist keypoints.

    \subsection{Multimodal Fusion}

        Multimodal fusion allows for obtaining a command by combining information from multiple unimodal communication channels that must be synchronized and merged together since different communication modalities have varying operating times and frequencies. For example, if the operator asks the robot "Bring me that object" using a "Point At" gesture, the gestural information is captured almost instantaneously and multiple times (at approximately 15 frames per second), while the vocal information must wait for the operator to complete the sentence before being processed. Therefore, it is necessary to pre-process the information through a \textbf{time manager}, whose task is to synchronize and combine them into a tensor that is then passed through a neural network classifier, which performs the multimodal fusion.
        The time manager, inspired by the recognition lines discussed in Cacace et al. \cite{cacace2017roman}, handles these delays and repetitions to combine and synchronize information related to the same command by receiving information from each channel for a specific \textit{time window} and encoding them into a tensor to be passed to the neural network. The time window opens when the first information is received and has an arbitrarily chosen duration. 
        The tensor synchronized by the time manager is then passed to a \textit{neural classifier} responsible for multimodal fusion, along with any additional information generated by the \textit{raw functions} if necessary. 
        The network, trained with a dataset created by collecting possible outcomes of the task, produces an output that represents a single multimodal command. This command can be an instruction to be given to the robot, a signal of error or lack of information to achieve a complete meaning, or a response to be provided to the operator using the text-to-speech feedback channel.

        \noindent
        \noindent

    \newpage
\section{Safety Layer}


    Once the message is received and forwarded to the robot, it is crucial for the robot to perform the desired task in a secure and efficient manner. 
    To accomplish this, the overall framework incorporates a well-defined motion planning strategy, called safety layer, responsible for planning trajectories that are safe for the human operator \cite{pupa2021safety}. 
    
    The implemented strategy operates in two stages. Initially, it computes a collision-free trajectory $\traj$, allowing the robot to ideally execute it at maximum speed. Subsequently, it dynamically adjusts the velocity along the path in real-time to ensure safety. This is achieved by employing a path-velocity decomposition technique, see equations \eqref{eq:q_s}-\eqref{eq:q_s_dot}, and solving online the following optimization problem:
    \begin{equation}
        \label{eq:problem}
        \begin{array}{ll@{}ll}
            \displaystyle \min_{\alpha} & -\alpha,                                                                                                                      \\[0.3cm]
            \text{s.t.}                                                                                                                                               \\[0.3cm]
            & \displaystyle J_{r_i}({\q})\dqs\dot{s}\alpha \leq v_{max_i}                                 &  & \forall i \in \{1,\dots,n\}, \\[0.3cm]
            & \displaystyle \dq_{min} \leq \dqs\dot{s}\alpha \leq \dq_{max},                                                                \\[0.3cm]
            & \displaystyle \ddot{\q}_{min}\leq \frac{\dqs\dot{s}\alpha - \dq}{T_r} \leq \ddot{\q}_{max},                                   \\[0.3cm]
            & \displaystyle 0 \leq \alpha \leq 1.                                                                                           \\[0.3cm]
        \end{array}
    \end{equation}
    $\alpha \in \begin{bmatrix}0, 1 \end{bmatrix}$ is the optimization variable and represents the scaling factor.  $J_{r_i}(\q)\in \mathbb{R}^{1 \times n}$ is a \textit{modified jacobian} that takes into account only the scalar velocity of the $i$-th link towards the human operator. $v_{max_i}$ is the velocity limit imposed by the ISO/TS 15066 \cite{isots}. 
    $\dq_{min} \in \mathbb{R}^{n}$ and $\dq_{max} \in \mathbb{R}^{n}$ are the joint velocity lower bounds and the joint velocity upper bounds, respectively. While $\ddot{\q}_{min} \in \mathbb{R}^{n}$
    and $\ddot{\q}_{max} \in \mathbb{R}^{n}$ are the acceleration limits. $\dq \in \mathbb{R}^{n}$ is the actual robot velocity and $T_r$ is the robot execution time.
    
    \vspace{\baselineskip}\noindent
    The goal of the optimization problem \eqref{eq:problem} is to maximize the scaling factor moving exactly at the planned velocity,  i.e. $\alpha = 1$ that is the maximum speed. However, when the human approaches the robot, the safety standards require to decrease the velocity until, in the worst case, stopping the robot. This is guaranteed by the solution $\alpha = 0$.
    
    \newpage
\section{Experimental Validation}


    The experimental validation\footnote{Video of experiments available at \href{https://doi.org/10.5281/zenodo.8083948}{https://doi.org/10.5281/zenodo.8083948} \cite{davide_ferrari_2023_8083948}} was carried out through a comparative experiment, simulating a daily task in a home environment.
    The objective of the experiment was to assist 
    a person in gathering items from the pantry to prepare a meal, using a collaborative manipulator to perform pick-and-place tasks. Communications were provided through the multimodal fusion architecture, utilizing both vocal and gesture channels to instruct the robot about which object it should pick. The first experiment was conducted without the safety layer, while it was enabled during the second experiment to compare the results obtained and highlight the differences and issues that may arise when disregarding safety regulations.
    

    \subsection{Implementation Details}

        The architecture was built using ROS \cite{ros}, dividing the various components into independent nodes in order to ensure the modularity of the architecture and make it compatible with multiple communication channels and multimodal fusion algorithms.
     
        \myparagraph{A. Voice Communication Channel}
        
            \noindent
            The voice communication channel has been built by developing an Amazon Alexa custom skill, which is an application that enables the voice assistant to perform customized tasks or provide information on specific topics. The skill consists of a \textit{front-end} that contains a set of customized \textit{intents} (request-response structures) that are connected to a \textit{back-end} responsible for gathering information or performing the requested tasks. The back-end was developed to run locally and integrated within ROS, using \textit{Flask Ask}, a Flask \cite{grinberg2018flask} extension that allows the creation of skills in Python, and \textit{ngrok} to expose the back-end and connect it to the front-end through HTTPS tunneling.

            \begin{figure}[htbp]
                \centering
                \includegraphics[trim={0 7cm 0 3cm}, clip, width=10cm]{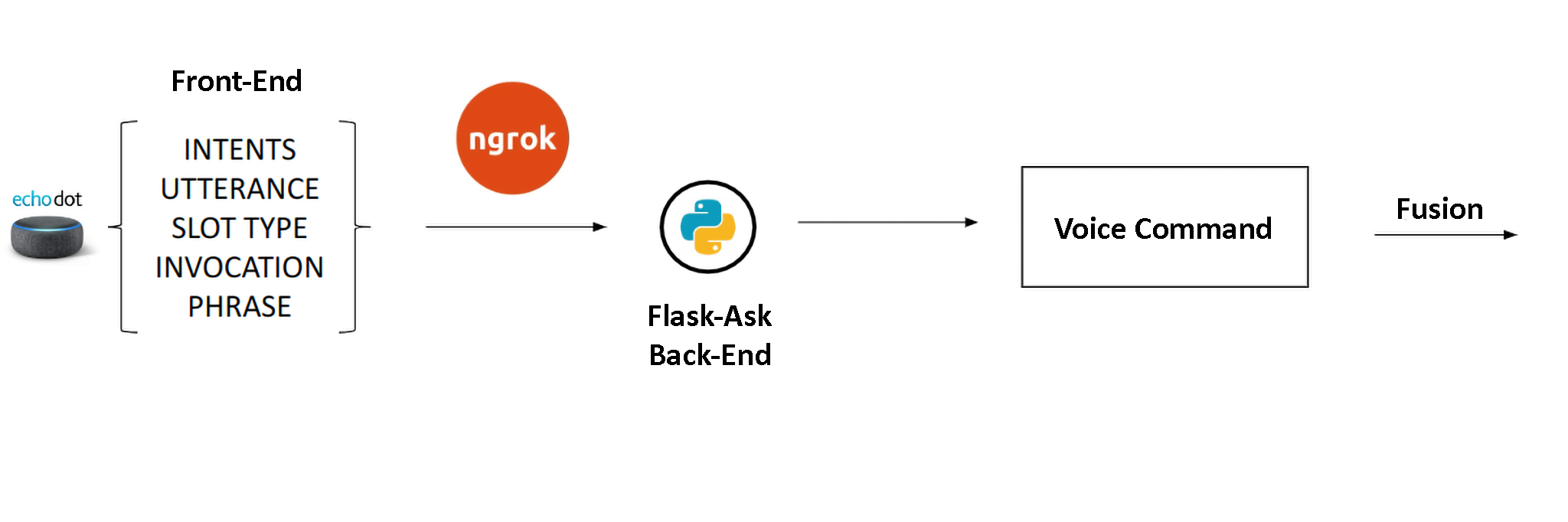}
                \caption{Voice Communication Channel}
                \label{fig:Voice Communication Channel}
            \end{figure}
    
            \noindent
            The Text-To-Speech has been created by integrating \textit{Node-RED} with ROS using the 'node-red-contrib-ros' node package. These nodes leverage the JavaScript library 'roslibjs' and connect to a ROS bridge through a WebSocket. The Text-to-Speech node enables the conversion of a text string into speech output, which is played on the Echo Dot device.

        \myparagraph{B. Gesture Recognition Channel}
        
            \noindent
            The gesture recognition channel was created using the Holistic landmarks detection solution API from MediaPipe \cite{lugaresi2019mediapipe}, a framework developed by Google that provides a suite of libraries and tools for applying artificial intelligence (AI) and machine learning (ML) techniques. Holistic combines components of pose, face, and hand landmarkers to create a comprehensive landmarker for the human body on a continuous stream of images in real-time.
        
            \begin{figure}[htbp]
                \centering
                \includegraphics[width=8cm]{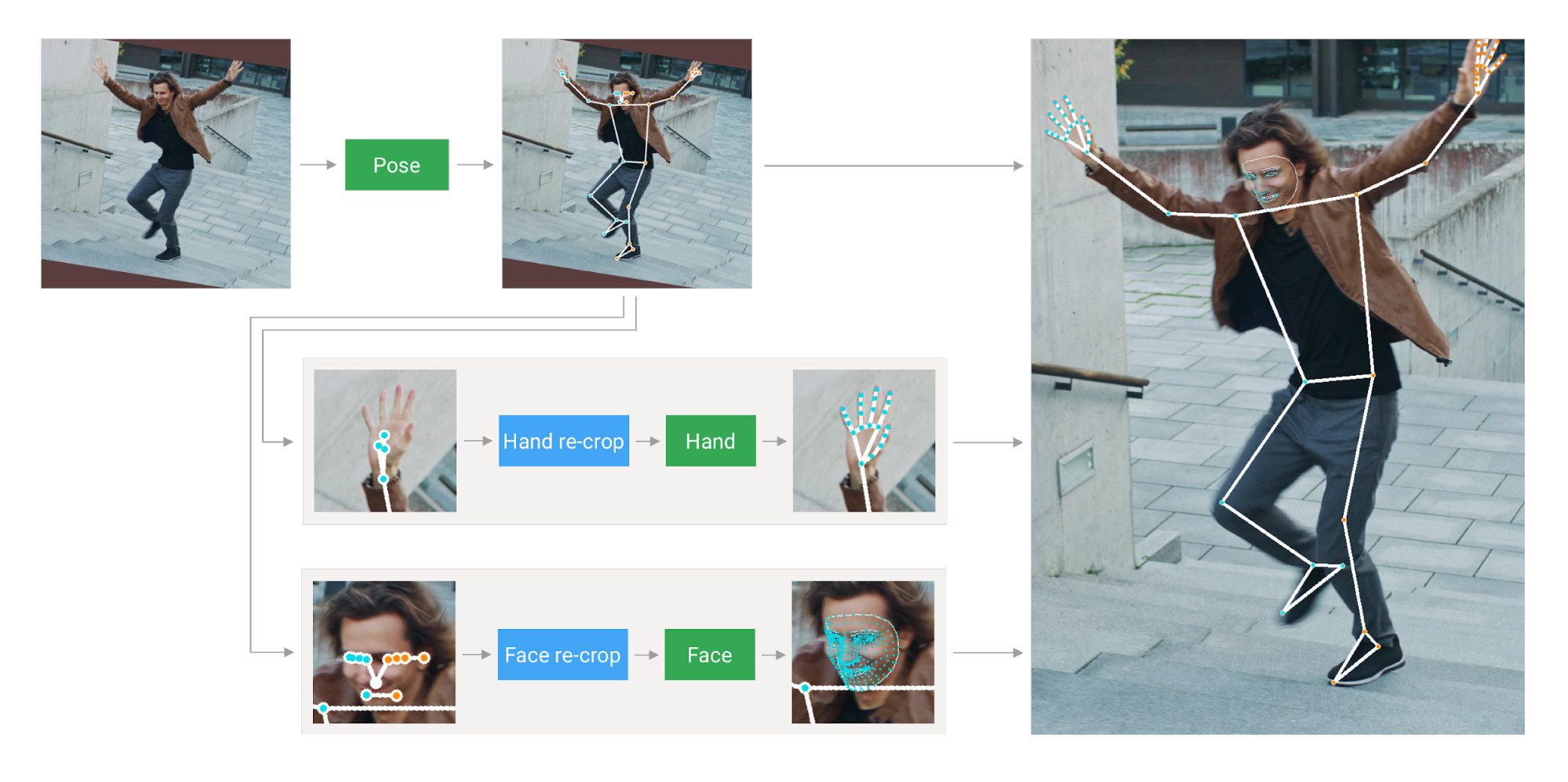}
                \caption{MediaPipe Holistic}
                \label{fig:MediaPipe Holistic}
            \end{figure}

            \noindent
            The landmarks extrapolated from each image are encoded into a tensor to represent the 3D gesture and then passed to a neural network classifier composed by an LSTM (Long Short-term Memory) layer followed by several fully connected layers, to classify gestures based on the landmarks. The neural network model was trained with a dataset of communicative gestures for human-robot collaboration, based on the work by Tan et al. \cite{tan2021proposed}.
            For each gesture, a series of videos were recorded and processed using MediaPipe Holistic to extract frame-by-frame landmarks, which were then encoded into matrices or tensors representing all the recordings of that specific gesture. These matrices of tensors were used for training the neural network model.
            
            \begin{figure}[htbp]
                \centering
                \includegraphics[width=10cm]{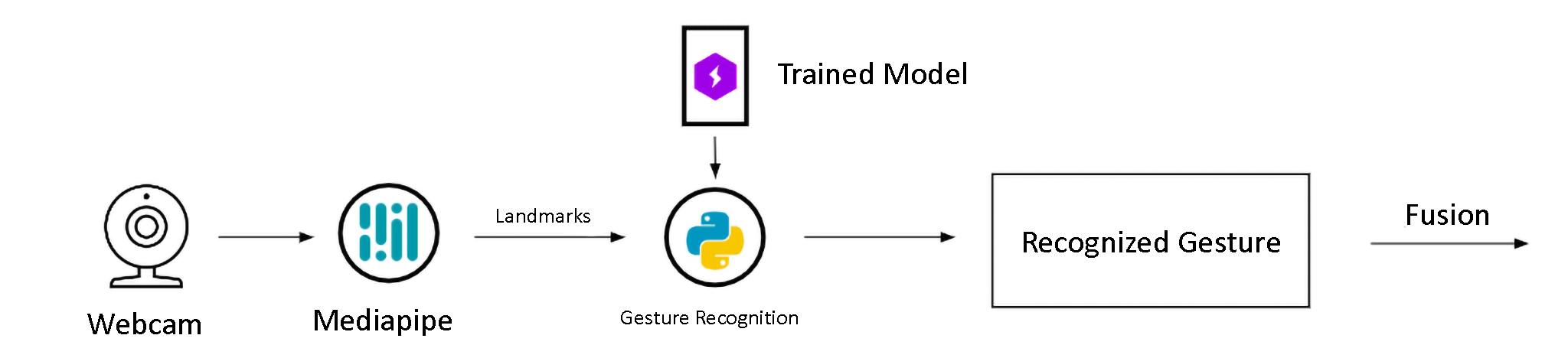}
                \caption{Gesture Recognition}
                \label{fig:Gesture Recognition}
            \end{figure}

    \newpage

    \myparagraph{C. Multimodal Fusion}

        \noindent
        The multimodal fusion is implemented using a time manager to aggregate and synchronize the flow of information into a tensor, that represents unimodal information pertaining to the same command, and a classifier neural network that takes the aforementioned tensor as input and outputs the fused multimodal command. As described in Algorithm \ref{algo:Multimodal-Fusion}, upon the arrival of new information on either the voice or gesture channel, a new temporal window $\mathcal{W}$ is opened, and the information is encoded into a tensor $\mathcal{T}$. During the duration of the temporal window, any new received information is appended to the tensor. Afterwards, the tensor is passed through a pre-trained neural network $\mathcal{N}$, which outputs the fused multimodal command $\mathcal{M}$.

        \begin{algorithm}[H]
            \begin{algorithmic}
                \\
                \Require Vectors $G$ and $V$ containing \textit{Gesture} and \textit{Voice} information
                \Ensure Multimodal Command $\mathcal{M}$
                \\
                \State $\mathcal{T} \gets$ empty tensor
                \State Recognition Time: $R_T \gets 2s$
                \\
                \While {new value of $G$ or $V$ is received}
                    \\
                    \State Open a \textit{Temporal Window} $\mathcal{W}$
                    \State $\mathcal{T} \gets$ Received Value ($G$ or $V$)
                    \\
                    \While {$\mathcal{W} < R_T$}
                        \\
                        \If {new gesture/voice $G$ or $V$ is received}
                            \State Append new $G$ or $V$ to tensor $\mathcal{T}$
                        \EndIf
                        \\
                    \EndWhile
                    \\
                    \State Pass tensor $\mathcal{T}$ through Neural Classifier $\mathcal{N}$ $\rightarrow$ Multimodal Command $\mathcal{M}$
                    \State Send the Multimodal Command $\mathcal{M}$ to the Safety Layer
                    \\
                \EndWhile
            \end{algorithmic}
            \caption{Multimodal Fusion Algorithm}
            \label{algo:Multimodal-Fusion}
        \end{algorithm}
        

    \myparagraph{D. Safety Layer}

        \noindent
        The safety layer is implemented exploiting \textit{fmincon} Matlab solver and works at the frequency of the robot controller, i.e. 500 Hz for the UR10e which has been used in this experiment. 
        The monitoring of the human operator in the scene is achieved exploiting 6 Optitrack Prime$^x$ cameras with the Motive software, which works at 240 Hz. 

    \newpage

    \subsection{Experiment Description and Results}

\begin{figure}[htbp]
            \minipage{0.49\textwidth}
                \includegraphics[trim={12cm 2.5cm 6cm 4.5cm}, clip, width=\linewidth]{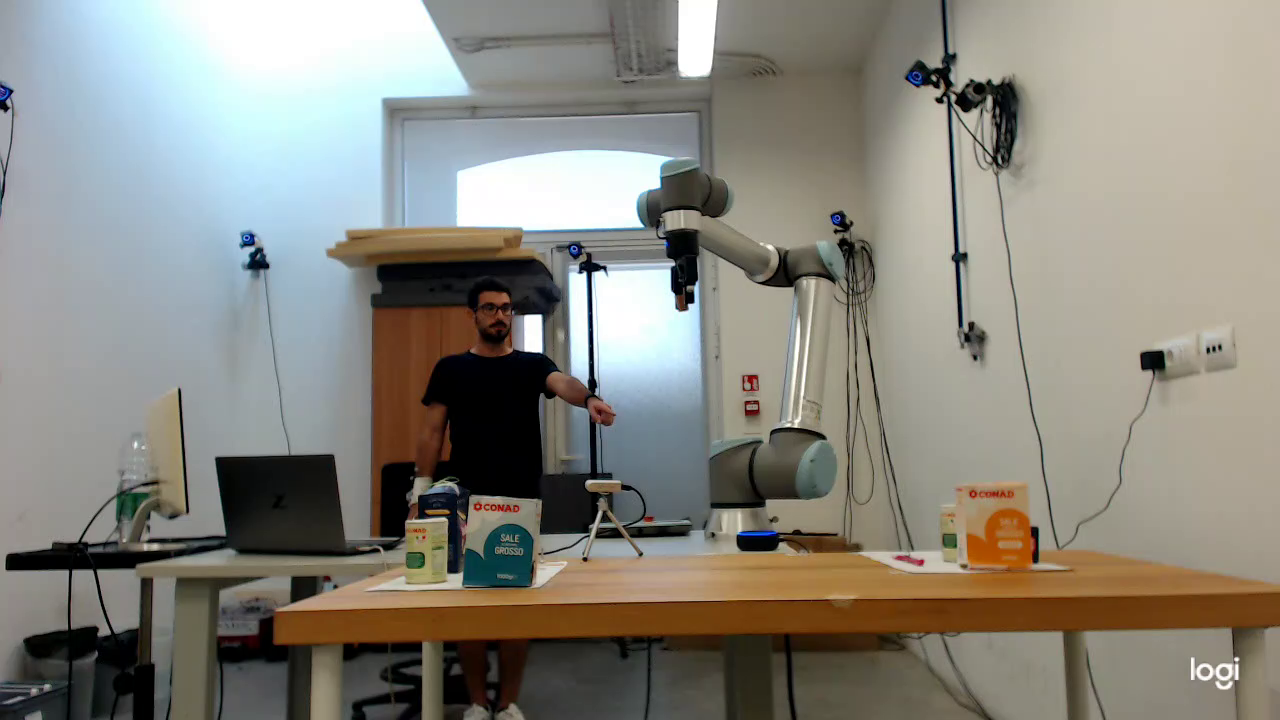}
            \endminipage\hfill
            \minipage{0.49\textwidth}
                \includegraphics[trim={0 1cm 0 2.3cm}, clip, width=\linewidth]{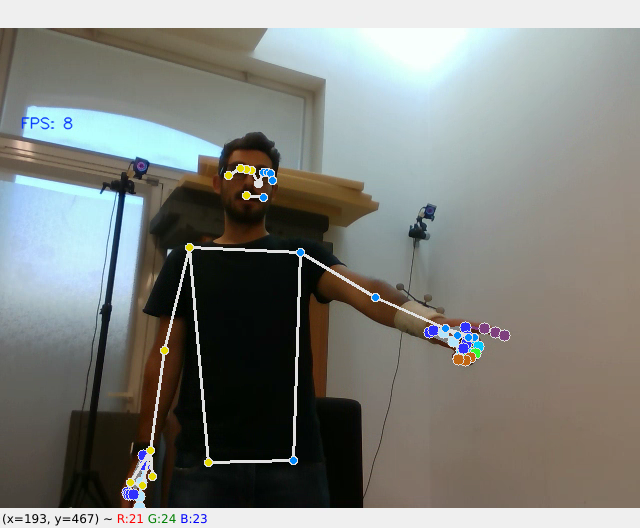}
            \endminipage\hfill
            \minipage{0.49\textwidth}
                \includegraphics[trim={6cm 2.5cm 6cm 4.5cm}, clip, width=\linewidth]{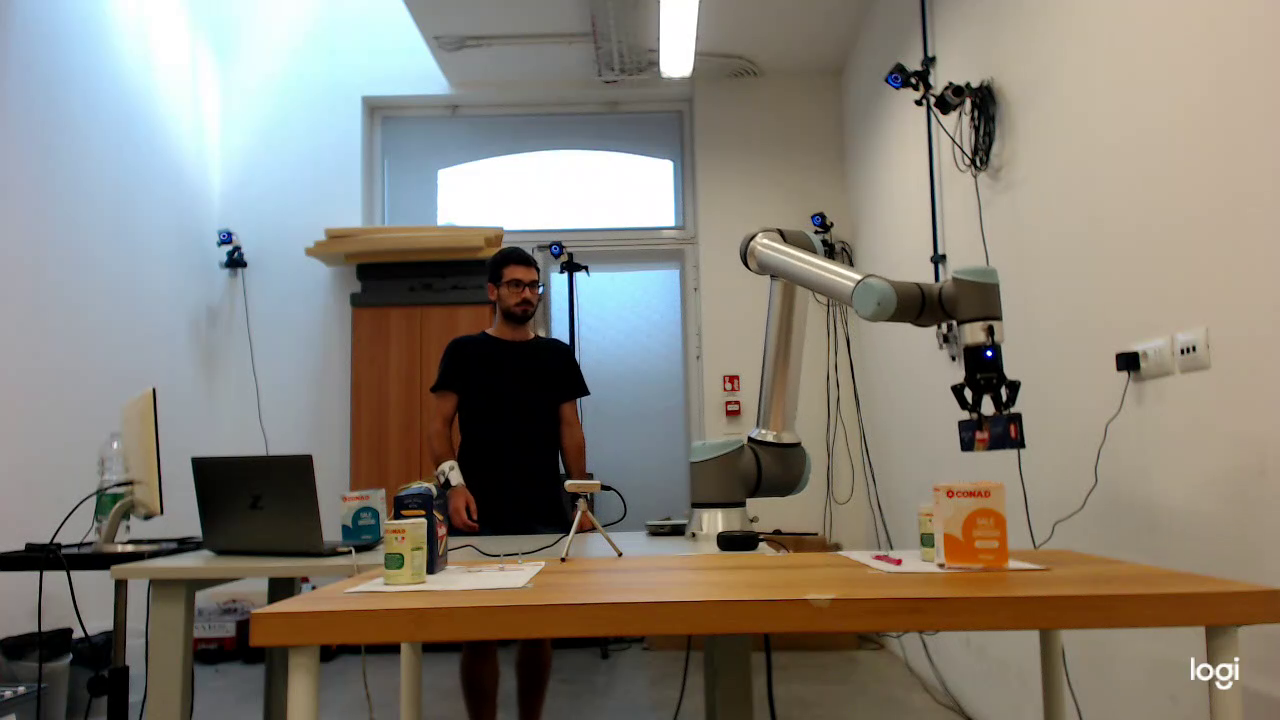}
            \endminipage\hfill
            \minipage{0.49\textwidth}
                \includegraphics[trim={6cm 2.5cm 6cm 4.5cm}, clip, width=\linewidth]{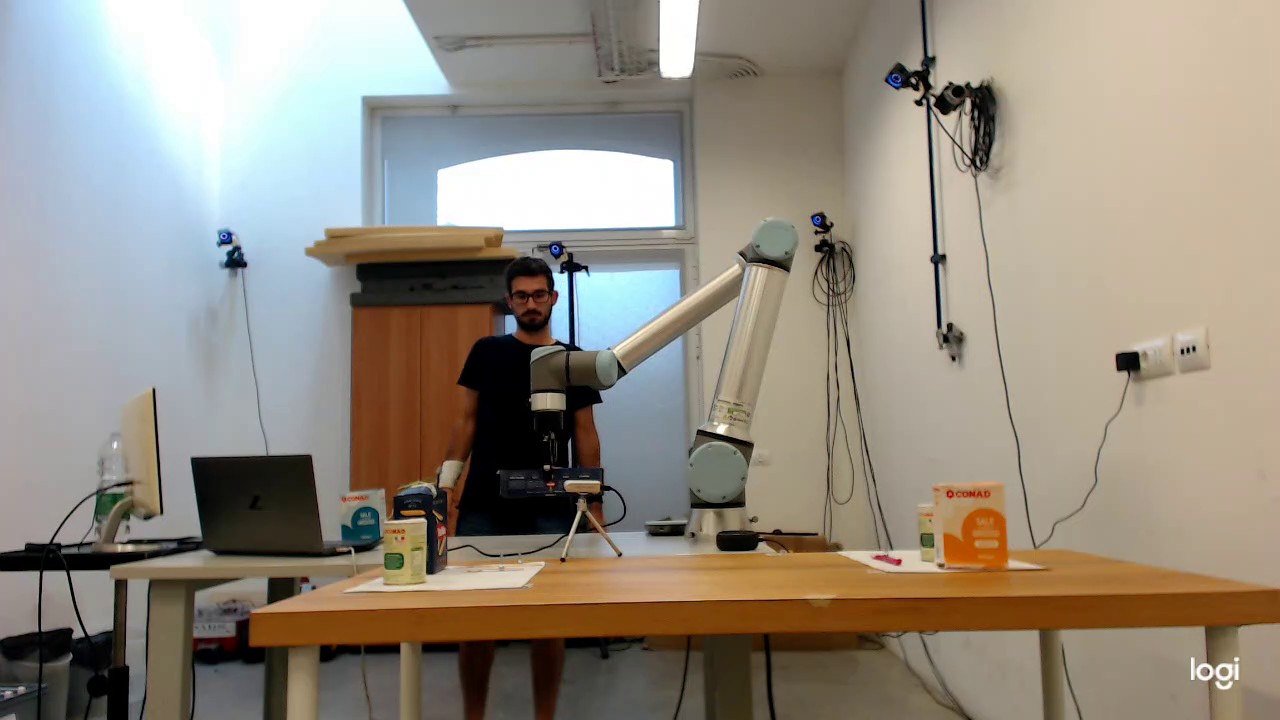}
            \endminipage\hfill
            \caption{Multimodal Fusion Object Request}
            \label{fig:Multimodal Fusion Object Request}
        \end{figure}

        The experiment consists of a series of pick-and-place tasks where the operator can request an object by describing it and indicating the corresponding area either through the voice channel alone (e.g., \textit{"Fetch me the pasta in the right area"}) or by combining the vocal command \textit{"bring me me the pasta"} with the gesture command \textit{"Point-At"} indicating the area. Figure \ref{fig:Multimodal Fusion Object Request} shows the operator requesting an object while indicating the corresponding area. By using the "raw functions", it is possible to calculate the direction in which the operator is pointing by interpolating the shoulder, elbow, and wrist landmarks. Consequently, this allows the multimodal fusion to obtain a complete command such as \textit{"take object x in the right/left area"}.

\begin{figure}[htbp]
            \centering
            \includegraphics[trim={2cm 2cm 2cm 2cm}, clip, width=12cm]{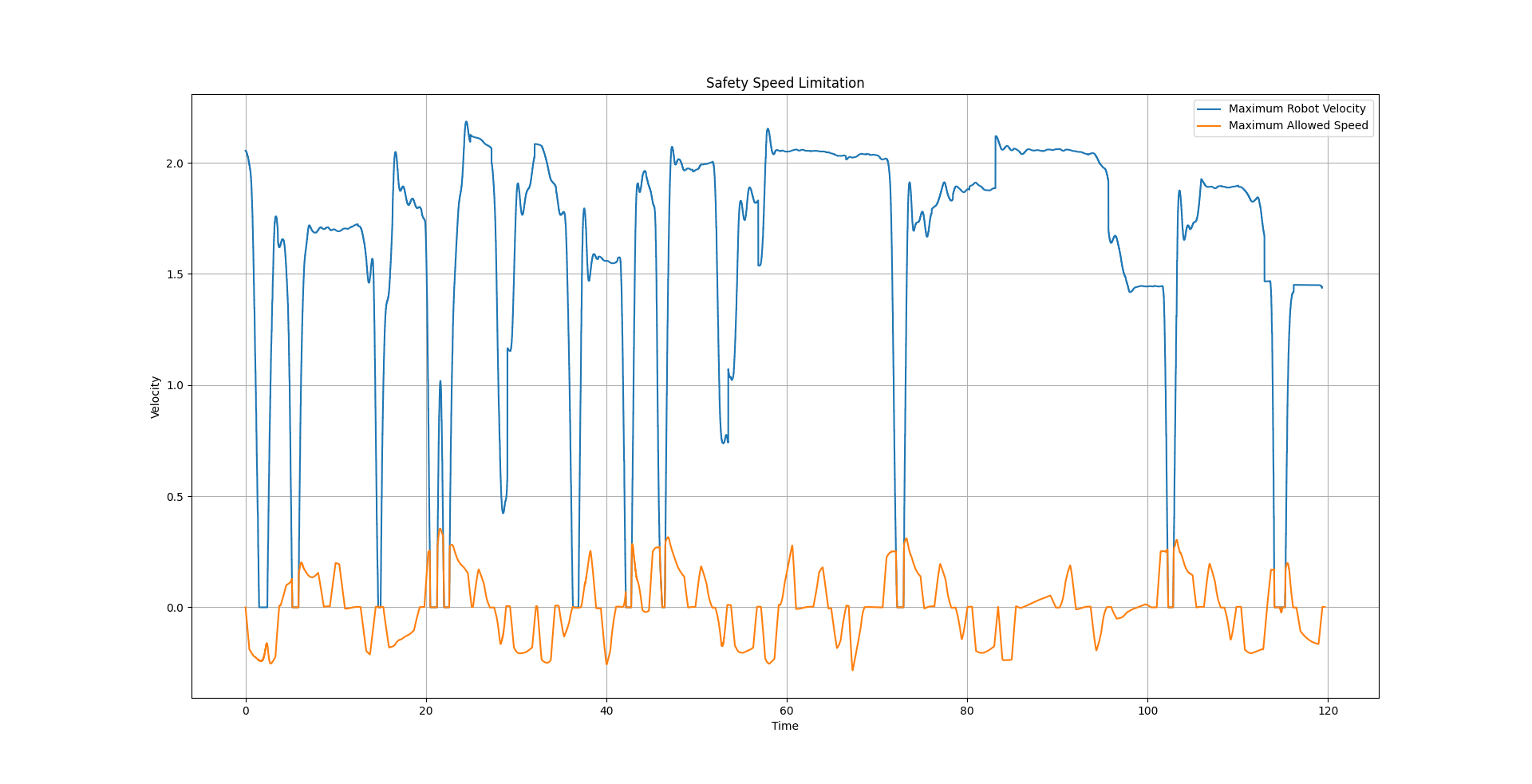}
            \caption{Maximum Allowed Velocity in relation to ISO Velocity Limit}
            \label{fig:safety}
        \end{figure}

        \noindent
        Multimodal Fusion combines the received information and sends the command to the safety layer, which is responsible for planning the trajectory and monitoring the operator to perform movements that are always safe and compliant with the ISO standards.
        In Figure \ref{fig:safety}, we can see how during the execution of the trajectory, the maximum speed of the manipulator is always below the speed limit required by the ISO. This speed limit is calculated based on the minimum distance from the operator and the direction of the robot's movement. If the robot is moving away from the human, the speed will be scaled down to ensure safety. However, if the robot is moving in a direction that does not pose any risk, it can operate at full speed.
        
\begin{figure}[htbp]
            \minipage{0.49\textwidth}
                \includegraphics[trim={6cm 2.5cm 6cm 4.5cm}, clip, width=\linewidth]{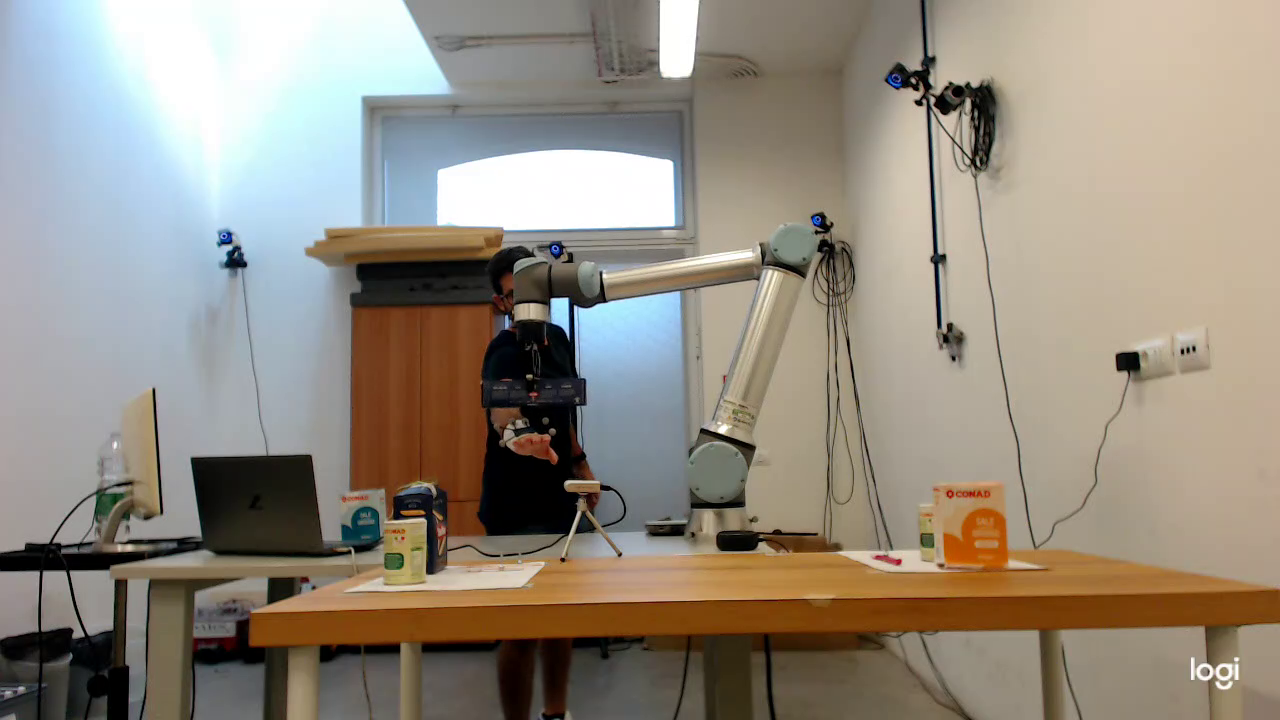}
            \endminipage\hfill
            \minipage{0.49\textwidth}
                \includegraphics[trim={6cm 2.5cm 6cm 4.5cm}, clip, width=\linewidth]{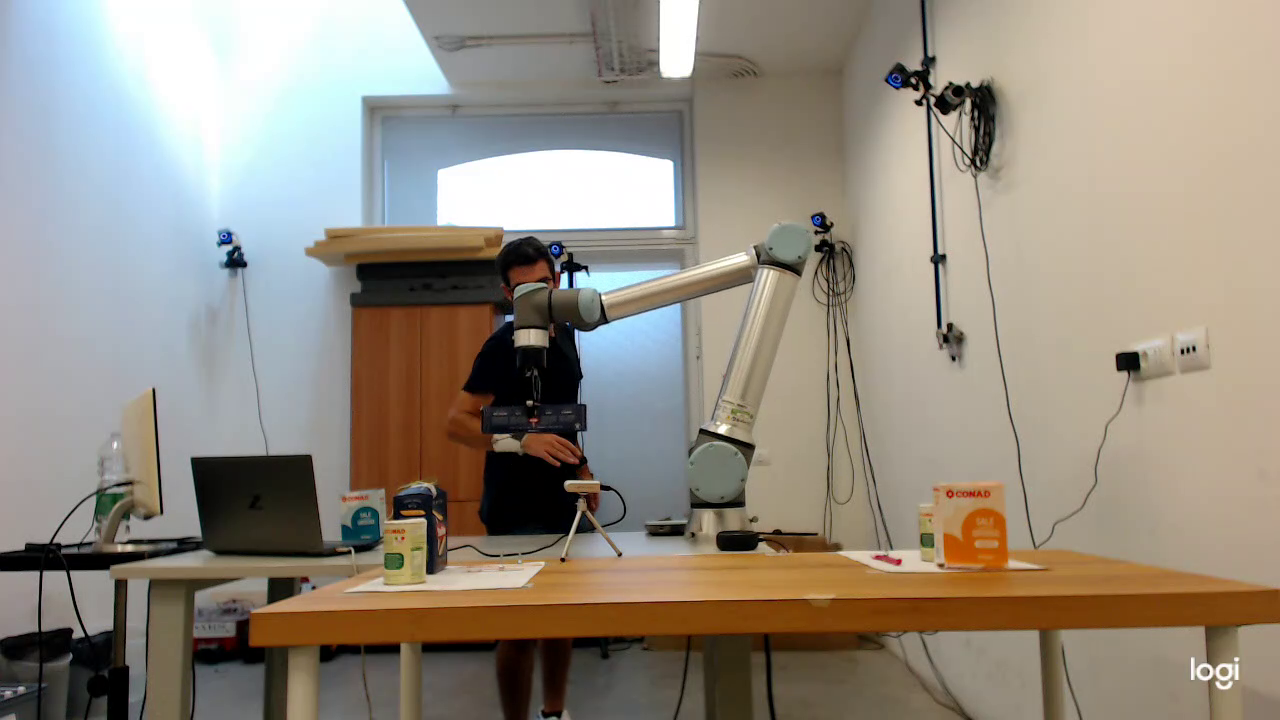}
            \endminipage\hfill
            \minipage{0.49\textwidth}
                \includegraphics[trim={6cm 2.5cm 6cm 4.5cm}, clip, width=\linewidth]{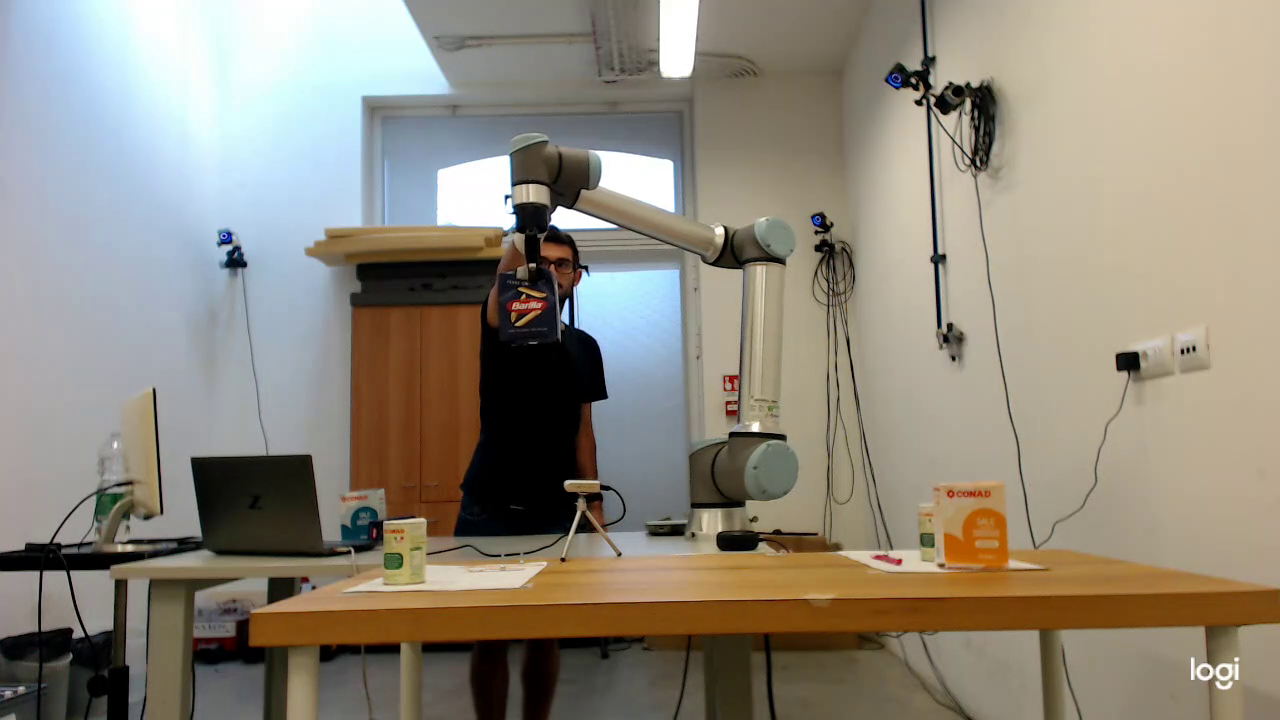}
            \endminipage\hfill
            \minipage{0.49\textwidth}
                \includegraphics[trim={6cm 2.5cm 6cm 4.5cm}, clip, width=\linewidth]{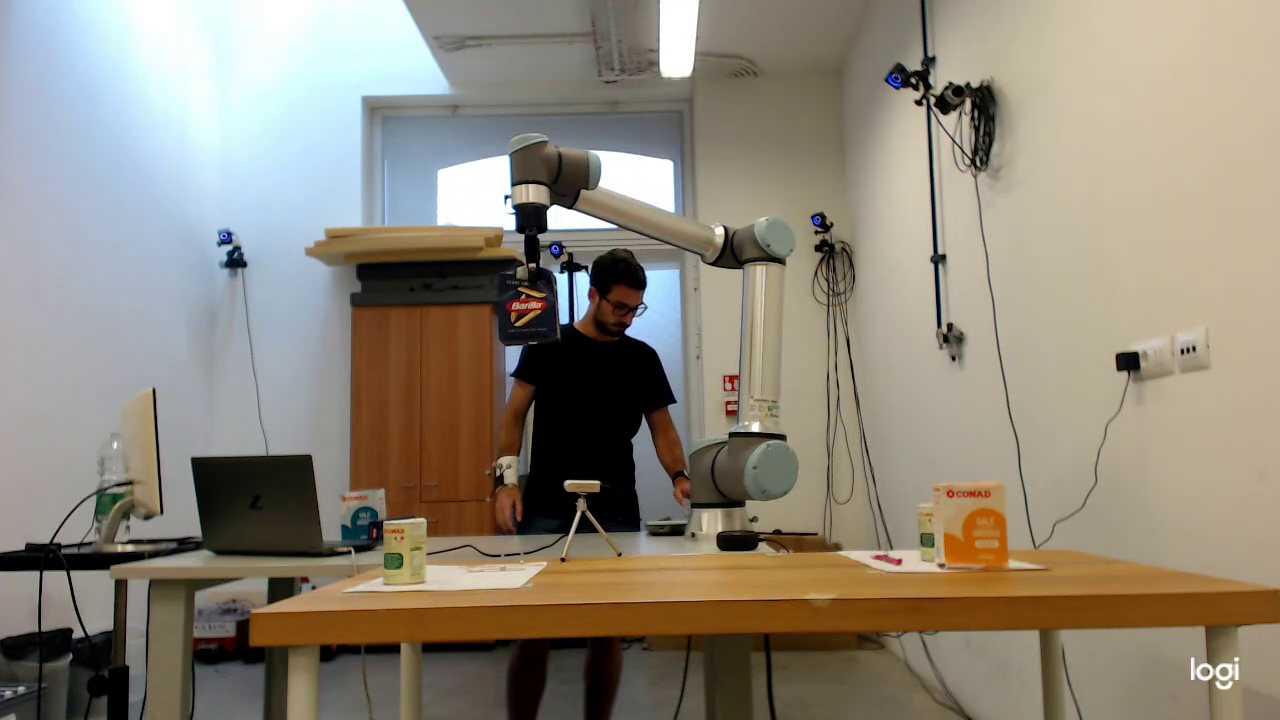}
            \endminipage\hfill
            \caption{Unsafe Experiment with Safety Layer Deactivated}
            \label{fig:Unsafe Behavior}
        \end{figure}

        \noindent
        On the contrary, when the safety is deactivated, the robot is not aware of the operator's position and performs unsafe trajectories that force the user to retract their arm to avoid a collision (Figure \ref{fig:Unsafe Behavior}). In more critical cases where the collision cannot be avoided, the robot triggers an emergency stop, requiring the operator to manually reinitialize the control algorithm.

    \section{Conclusions}

    This paper presented a multimodal communication architecture that integrates voice and gestures to achieve a simpler and more natural interaction with the robot, with particular attention to ensuring safety during collaboration. 
    To validate its effectiveness, we conducted a comparative experiment, with and without the safety layer, simulating a daily task in a home environment. This experiment confirmed the ability of the architecture to correctly fuse the operator's communications and successfully complete all the required tasks.
    Furthermore, we highlighted how the lack of attention to safety regulations can jeopardize the operator's safety during close collaboration with a robot. Our architecture, prioritizing safety at all times, enables a simple and natural interaction with the robot, avoiding situations of danger for the operator.
    As further extensions of this architecture, we evaluated the idea of adding more communication channels to improve and expand the ability of the system to interact with the operator. We have also planned to leverage the communication channels to collaborate with the safety layer in resolving any errors, such as obstacles in the path or alerts of critical situations where the robot needs to move in the area occupied by the operator. This approach aims to increase the complexity of the exchanged information, bringing us closer to achieving communication that is more similar to that between human beings.

    \bibliographystyle{Styles/bibtex/splncs03_unsrt}
    \bibliography{Bibliography.bib}

\end{document}